\documentclass[runningheads]{llncs}
\usepackage{graphicx}

\usepackage{tikz}
\usepackage{comment}
\usepackage{amsmath,amssymb}
\usepackage{color}
\usepackage[citecolor=green]{hyperref}

\usepackage{algorithm}
\usepackage{algorithmic}
\usepackage[accsupp]{axessibility} 
\usepackage{makecell} 
\usepackage{booktabs}

\begin{document}
\pagestyle{headings}
\mainmatter
\def\ECCVSubNumber{1382} 

\title{RCLane: Relay Chain Prediction for Lane Detection}

\titlerunning{RCLane: Relay Chain Prediction for Lane Detection}
\author{
Shenghua Xu\inst{1}\thanks{
Equal contribution.}
\and 
Xinyue Cai\inst{2}$^\star$
\and 
Bin Zhao\inst{1}
\and 
Li Zhang\inst{1}\thanks{Li Zhang (lizhangfd@fudan.edu.cn) is the corresponding author with School of Data Science, Fudan University.}
\and 
Hang Xu\inst{2}
\and \\
Yanwei Fu\inst{1}
\and 
Xiangyang Xue\inst{1}
}

\authorrunning{Shenghua Xu et al.}
\institute{Fudan University
\and Huawei Noah's Ark Lab}

\maketitle

\begin{abstract}
Lane detection is an important component of many real-world autonomous systems. 
Despite a wide variety of lane detection approaches have been proposed, reporting steady benchmark improvements over time, lane detection remains a largely unsolved problem.
This is because most of the existing lane detection methods either treat the lane detection as a dense prediction or a detection task, few of them consider the unique topologies (\textit{Y-shape, Fork-shape, nearly horizontal lane}) of the lane markers, which leads to sub-optimal solution.
In this paper, we present a new method for lane detection based on \textit{relay chain} prediction.
Specifically, our model predicts a segmentation map to classify the foreground and background region.
For each pixel point in the foreground region, we go through the forward branch and backward branch to recover the whole lane.
Each branch decodes a transfer map and a distance map to produce the direction moving to the next point, and how many steps to progressively predict a relay station (next point).
As such, our model is able to capture the keypoints along the lanes.
Despite its simplicity, our strategy allows us to establish new state-of-the-art on four major benchmarks including \textit{TuSimple, CULane, CurveLanes} and \textit{LLAMAS}.

\keywords{Lane detection, relay chain.}
\end{abstract}
\section{Introduction}

Lane detection, the process of identifying lanes as approximated curves, is a fundamental step in developing advanced autonomous driving system and plays a vital role in applications such as driving route planning, lane keeping, real-time positioning and adaptive cruise control.
\begin{figure}[ht]
    \setlength{\abovecaptionskip}{0.1cm}
    \centering
    \includegraphics[width=1\linewidth]{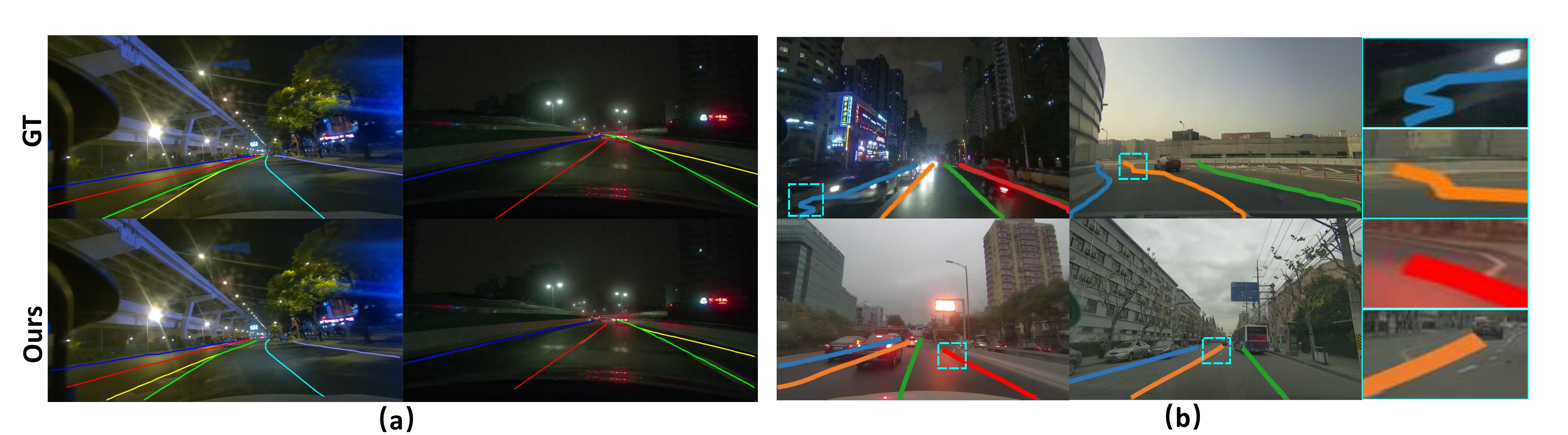}
    \caption{\textbf{Challenging scenes (curve lanes, Y-shape lanes).} The first row of (a) shows the ground truth while the second row is our predictions. The first row of (b) shows the result of segmentation-based methods that global shape of lane is not well fitted. While the second row of (b) shows proposal-based methods, can not depict local locations of Y-shape and curve lanes.}
    \label{fig:teaser}
\end{figure}

Early lane detection methods~\cite{liu2010combining,tan2014novel,zhou2010novel,borkar2009robust,hur2013multi,jiang2009new,jiang2010computer,kim2008robust} usually extract hand-crafted features and cluster foreground points on lanes through post-processing. 
However, traditional methods can not detect diverse lanes correctly for so many complicated scenes in driving scenarios. 
Thanks to the development of deep-learning, a wide variety of lane detection approaches based on convolution neural network(CNN) have been proposed, such as segmentation-based methods and proposal-based methods, reporting steady benchmark improvements over time. 

Proposal-based methods initialize a fixed number of anchors directly and model global information  focusing on the optimization of proposal coordinates regression. 
LaneATT~\cite{Tabelini2020KeepYE} designs slender anchors according to long and thin characteristic of lanes. 
However, line proposals fail to generalize local locations of all lane points for curve lanes or lanes with more complex topologies.
While segmentation-based methods treat lane detection as dense prediction tasks to capture local location information of lanes. 
LaneAF~\cite{abualsaud2021laneaf} focuses on local geometry to integrate into global results. 
However, this bottom-up manner can not capture the global geometry of lanes directly. 
In some cases such as occlusion or resolution reduction for points on the far side of lane, model performance will be affected due to the loss of lane shape information.
Visualization results in Fig.~\ref{fig:teaser}(b) of these methods show their shortcomings.
Lanes always span half or almost all of the image, these methods neglect this long and thin characteristic of lanes which requires networks to focus on the global shape message and local location information simultaneously. 
In addition, complex lanes such as Y-shape lanes and Fork-shape lanes are common in the current autonomous driving scenario, while existing methods often fail at these challenging scenes which are shown in Fig.~\ref{fig:teaser}(a).

To address this important limitation of current algorithms, we propose a more accurate lane detection solution in the unconstrained driving scenarios, which is called \textit{RCLane} inspired by the idea of \textbf{\textit{Relay Chain}} for focusing on local location and global shape information of lanes at the meanwhile.
Each foreground point on the lane can be treated as a relay station for recovering the whole lane sequentially in a chain mode.
Relay station construction is proposed for strengthening the model's ability of learning local message that is fundamental to describe flexible shapes of lanes. 
To be specific, we construct a transfer map representing the relative location from current pixel to its two neighbors on the same lane.
Furthermore, we apply bilateral prediction strategy aiming to improve generalization ability for lanes with complex topologies.
Finally, we design global shape message learning module.
Concretely, this module predicts the distance map describing the distance from each foreground point to the two end points on the same lane. 
The contributions of this work are as follows: 
\begin{itemize}
    \item We propose novel relay chain representation for lanes to model global geometry shape and local location information of lanes simultaneously.
    \item We introduce a novel pair of lane encoding and decoding algorithms to facilitate the process of lane detection with relay chain representation.
    \item  Extensive experiments on four major lane detection benchmarks show that our approach beats the state-of-the-art alternatives, often by a clear margin and achieves real-time performance.
\end{itemize}

\section{Related work}
Existing methods for lane detection can be categorized into:
segmentation-based methods, proposal-based methods, row-wise methods and polynomial regression methods.

\noindent\textbf{Segmentation-based methods.}
Segmentation-based methods \cite{Hou2019LearningLL,Ko2020KeyPE,Lee2017VPGNetVP,Neven2018TowardsEL,Pan2018SpatialAD}, typically make predictions based on pixel-wise classification. 
Each pixel will be classified as either on lane or background to generate a binary segmentation mask. 
Then a post-processing step is used to decode it into a set of lanes. 
But it is still challenging to assign different points to their corresponding lane instances. 
A common solution is to predict the instance segmentation mask. 
However, the number of lanes has to be predefined and fixed when using this strategy, which is not robust for real driving scenarios.

\noindent\textbf{Proposal-based methods.}
Proposal-based methods~\cite{8813778,Xu2020CurveLaneNASUL,Tabelini2020KeepYE},
take a top-to-down pipeline that directly regresses the relative coordinates of lane shapes.
Nevertheless, they always struggle in lanes with complex topologies such as curve lanes and Y-shaped lanes. 
The fixed anchor shape has a major flaw when regressing the variable lane shapes in some hard scenes.

\noindent\textbf{Row-wise methods.}
Based on the grid division of the input image, row-wise detection approaches~\cite{Hou2020InterRegionAD,Philion2019FastDrawAT,Qin2020UltraFS,Yoo2020EndtoEndLM,Liu2021CondLaneNetAT} have achieved great progress in terms of accuracy and efficiency.
Generally, row-wise detection methods directly predict the lane position for each row and construct the set of lanes through post-processing. 
However, detecting nearly horizontal lanes which fall at small vertical intervals is still a major problem. 

\noindent\textbf{Polynomial regression methods.}
Polynomial regression methods~\cite{Liu2021EndtoendLS,Tabelini2021PolyLaneNetLE} directly outputs polynomials representing each lane.
The deep network is firstly used in~\cite{Tabelini2021PolyLaneNetLE} to predict the lane curve equation, along with the domains for these polynomials and confidence scores for each lane. 
\cite{Liu2021EndtoendLS} uses a transformer~\cite{Vaswani2017AttentionIA} to learn richer structures and context, and reframes the lane detection output as parameters of a lane shape model.
However, despite of the fast speed polynomial regression methods achieve, there is still some distance from the state of the art results.
\begin{figure*}[t]
\centering
\includegraphics[width=1\linewidth]{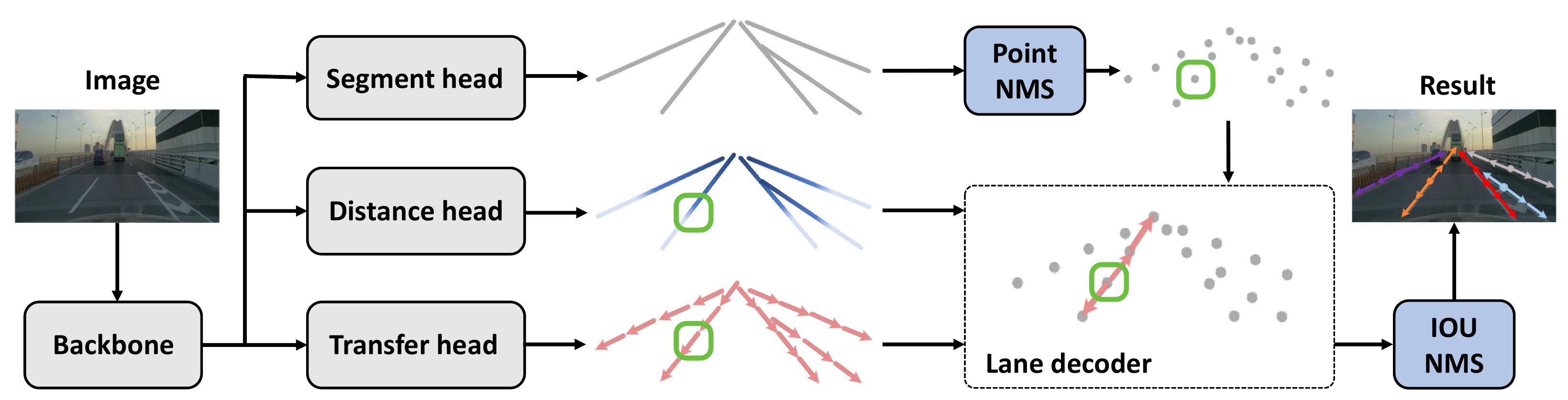}
\caption{\textbf{Schematic illustration of proposed RCLane.} Standard Segformer~\cite{Xie2021SegFormerSA} is used as backbone. The output head consists of three branches. The segment head predicts segmentation map (\textit{S}). The distance head and the transfer head predict distance map (\textit{D}) and transfer map (\textit{T}) respectively. Both kinds of maps contain forward and backward parts. Then, Point-NMS is used for sparse segmentation results. All predictions are fed into the lane decoder (Fig.~\ref{fig:Lane_decoder}), to get final results.}
\label{fig:network archi}
\end{figure*}
\section{Method}

Given an input image $I \in \mathbb{R}^{H \times W \times C}$, the goal of RCLane is to predict a collection of lanes $L = \{l_1, l_2, \cdots, l_N\}$, where $N$ is the total number of lanes. Generally, each lane $l_k$ is represented as follows:
\begin{equation}
l_k = \{(x_{1}, y_{1}), (x_{2}, y_{2}), \cdots, (x_{N_k}, y_{N_k})\},
\end{equation}

The overall structure of our RCLane is shown in Fig.~\ref{fig:network archi}. 
This section will first present the concept of lane detection with relay chain, then introduce the lane encoder for relay station construction, followed by a lane decoder to attain curve lanes. 
Finally, the network architecture and losses we adopt is detailed.

\subsection{Lane detection with relay chain}
Focusing on the combination of local location and global shape information to detect lanes with complex topologies, we propose a novel lane detection method RCLane with the idea of relay chain.
Relay chain is a structure composed of relay stations which are connected in a chain mode.
Relay station is responsible for data processing and transmitting it to adjacent stations, while chain is a kind of structure that organizes these stations from an overall perspective. 
All stations are associated to corresponding lane points respectively. 

We design the structure of relay chain which is appropriate for combining local location and global geometry message in lane detection and propose RCLane in this work.
To be specific, each foreground point on the lane is treated as a relay station and can extend to the neighbor points iteratively to decode the lane in a chain mode. 
All foreground points are supervised by two kinds of message mentioned above. 
Moreover, the structure of chain has high flexibility to fit lanes with complex topologies.

Next, we will introduce the relay station construction and propose bilateral predictions for complex topologies and global shape message learning to explain how to detect lanes with the idea of \textit{Relay Chain} progressively.

\noindent\textbf{Relay station construction.}
Segmentation-based approaches normally predict all foreground points on lanes and cluster them via post-processing. 
\cite{abualsaud2021laneaf} predicts horizontal and vertical affinity fields for clustering and associating pixels belonging to the same lane.
\cite{qu2021focus} regresses a vector describing the local geometry of the curve that current pixel belongs to and refines shape further in the decoding algorithm. 
Nevertheless, they both fix the vertical intervals between adjacent points and decode lanes row-by-row from bottom to top. 
In fact, horizontal offsets are used for refining the position of current points while vertical offsets are for exploring the vertical neighbors of them.
And the fixed vertical offsets can not adapt to the high degree of freedom for lanes.
For example, they can only detect a fraction of the nearly horizontal lanes. 
Thus, we propose relay station construction module to establish relationships between neighboring points on the lane. 
Each relay station $p=(p_x, p_y)$ predicts offsets to its neighboring point $p^{next}=(p^{next}_x, p^{next}_y)$ on the same lane with a fixed step length $d$ as is shown in Eq.~\ref{eq:neighbor},~\ref{eq:step_length} in two directions. 
And the deformation trend of lanes can be fitted considerably by eliminating vertical constraints. 
All relay stations are then connected to form a chain which is the lane exactly.

\begin{equation}
(p^{next}_x, p^{next}_y) = (p_x, p_y) + (\Delta x, \Delta y),
\label{eq:neighbor}
\end{equation}

\begin{equation}  
\Delta x^2 + \Delta y^2 = d^2.
\label{eq:step_length}
\end{equation}

\begin{figure}[t]
    \setlength{\abovecaptionskip}{0.1cm}
    \centering
    \includegraphics[width=1\linewidth]{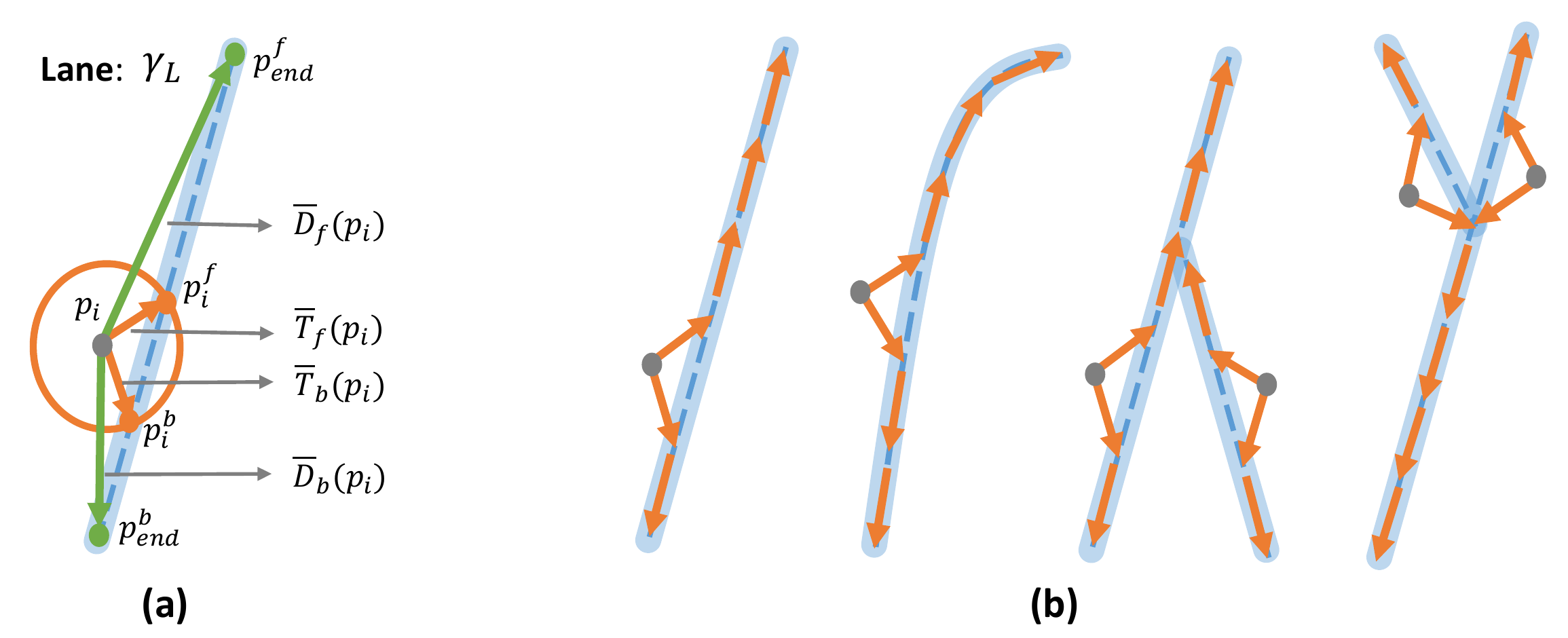}
    \caption{(a) is an illustration of the transfer vectors and distance scalars for $p_i$. $\overline{T}_{f,b}(p_i)$ are the forward and backward transfer vectors. $\overline{D}_{f,b}(p_i)$ are the forward and backward distance scalars. (b) shows our bilateral predictions can not only decode Y-shape or fork-shape lanes, but also fit simple structures, like straight lanes and curved lanes. }
    \label{fig:Encoder2decoder}
\end{figure}

\noindent\textbf{Bilateral predictions for complex topologies.}
The current autonomous driving scenario contains lanes with complex topologies such as Y-shape and  Fork-shape lanes, which can be regarded as that two lanes merges as the stem. 
One-way prediction can only detect one of lanes because it can only extend to one limb when starting from the stem of these lanes. 
We adopt a two-way detection strategy that splits the next neighboring point \textit{\protect $p^{next}$} into the forward point \textit{\protect $p^f$} and the backward point \textit{\protect $p^b$}. 
Points on different limbs can recover lanes they belong to respectively and compose the final Y-shape or fork-shape lanes as is illustrated in Fig.~\ref{fig:Encoder2decoder}(b).
Let \textit{F} denotes the output feature map from the backbone whose resolution drops by a factor of 4 compared to the original image. We design a transfer output head and pick \textit{F} as input. 
\textit{F} goes through convolution-based transfer head to get the transfer map \textit{T} which consists of forward and backward components $T_f, T_b \in \mathbb{R}^{H \times W \times 2}$. 
Each location in $T_f$ is a 2D vector, which represents the offsets between the forward neighboring point $p^f$ and the current pixel $p$. 
The definition of $T_b$ is similar as $T_f$. Consequently, we can detect the forward and backward neighboring points $p^f$, $p^b$ of \textit{p} guided by \textit{T}.

\begin{equation}
p^f = p + T_f(p),\quad {p^b} = p + T_b(p).
\end{equation}

With the guidance of local location information in transfer map \textit{T}, the whole lane can be detected iteratively via bilateral strategy.

\noindent\textbf{Global shape message learning.} 
Previous works predict positions of end points for lanes to guide decoding process.
FastDraw~\cite{Philion2019FastDrawAT} predicts end tokens to encode the global geometry while CondLaneNet~\cite{Liu2021CondLaneNetAT} recovers the row-wise shape through the vertical range prediction. 
These methods actually ignores the relation between the end points and other points on the same lane.
We make every relay station learns the global shape message transmitted in the chain by utilizing the relation mentioned above.
In detail, we design a distance head to predict the distance map \textit{D} that consists of the forward and backward components $D_f, D_b \in \mathbb{R}^{H \times W \times 1}$. 
Each location in $D_f$ is a scalar, which represents the distance from the current pixel $p$ to the forward end point $p_{end}^{f}$ on the lane. 
With this global shape information, we can know when to stop the lane decoding process. 
Specifically speaking, the iterations for decoding the forward branch of \textit{p} is \textit{\protect $\frac{D_f}{d}$}. 
The definition of $D_b$ is similar as $D_f$ as well. 
With the combination of local location and global geometry information, our relay chain prediction strategy performs considerably well even in complex scenarios. 
Next, we will introduce the novel pair of lane encoding and decoding algorithms designed for lane detection.

\subsection{Lane encoder for relay station construction}
The lane encoder is to create the supervision of transfer and distance maps for training.
Given an image $I \in \mathbb{R}^{H \times W \times 3}$ and its segmentation mask $\overline{S} \in \mathbb{R}^{H \times W \times 1}$, for any foreground point $p_i = (x_i, y_i) \in \overline{S}$ we denote its corresponding lane as $\gamma_L$. 
The two forward and backward end points of $\gamma_L$ are denoted as $p_{end}^{f} = (x_{end}^{f}, y_{end}^{f})$ and $p_{end}^{b} = (x_{end}^{b}, y_{end}^{b})$, which 
have the minimum and maximum y-coordinates respectively. 
The forward distance scalar $\overline{D}_f(p_i)$ and backward distance scalar $\overline{D}_b(p_i)$ of $p_i$ are formulated as the following:

\begin{equation}
\overline{D}_f(p_i) = \sqrt{(x_i-x_{end}^{f})^2 + 
(y_i-y_{end}^{f})^2},
\end{equation}

\begin{equation}
\overline{D}_b(p_i) = \sqrt{(x_i-x_{end}^{b})^2 + 
(y_i-y_{end}^{b})^2}.
\end{equation}

To generate the forward transfer vector and backward transfer vector for pixel $p_i$, we first find the two neighbors on $\gamma_L$ of it with the fixed distance $d$. 
They are denoted as $p_i^f=(x_i^{f}, y_i^{f})$ and $p_i^b=(x_i^{b}, y_i^{b})$ and represent the forward neighbor and backward neighbor respectively. 
Then the forward transfer vector $\overline{T}_f(p_i)$ and the backward transfer vector $\overline{T}_b(p_i)$ for pixel $p_i$ are defined :

\begin{equation}
\overline{T}_f(p_i) = (x_i^{f} - x_i, y_i^{f} - y_i),
\end{equation}

\begin{equation}
\overline{T}_b(p_i) = (x_i^{b} - x_i, y_i^{b} - y_i),
\end{equation}

\begin{equation}
\vert\vert \overline{T}_f(p_i) \vert\vert_2 = \vert\vert \overline{T}_b(p_i) \vert\vert_2  = d.
\end{equation}

\begin{figure*}[t]
    \setlength{\abovecaptionskip}{0.2cm}
    \centering
    \includegraphics[width=1\linewidth]{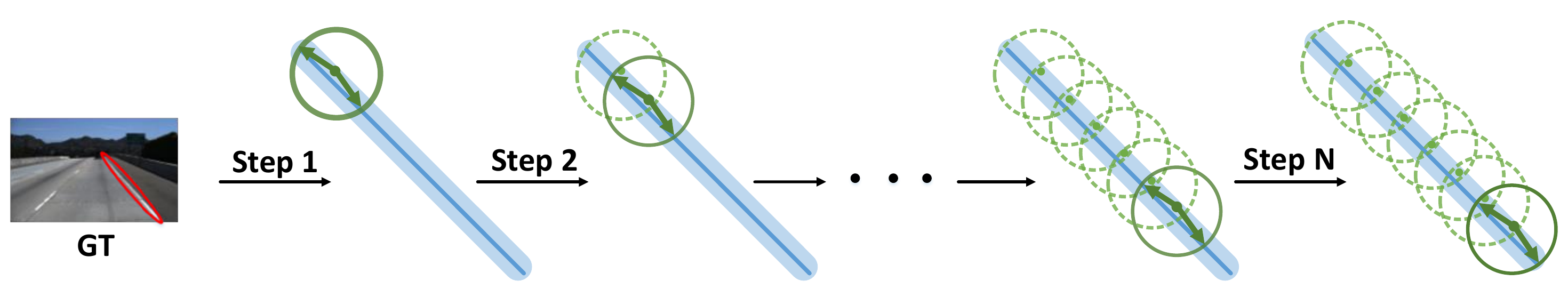}
    \caption{\textbf{Lane encoder.} All foreground points are matched with the nearest lanes. The arrows in a circle indicate transfer vectors of a foreground point to its two neighbors on lane. The distance scalars represent distances between the current point and two end points of the lane. All results are generated with point-wise traversal.}
    \label{fig:lane_encoder}
\end{figure*}

The details are shown in Fig.~\ref{fig:Encoder2decoder}(a). In addition, for two separate parts of one Y-shape lane:
$l_1 = \{(x_1, y_1), \cdots, (x_m, y_m),(x^1_{m+1}, y^1_{m+1}), \cdots, (x^1_{n_1}, y^1_{n_1})\}$, 
$l_2 = \{(x_1, y_1), \cdots, (x_m, y_m),(x^2_{m+1}, y^2_{m+1}), \cdots, (x^2_{n_2},y^2_{n_2})\}$. 
$\{(x_1, y_1), \cdots, (x_m, y_m)\}$ is the shared stem. 
We randomly choose one point from $(x^1_{m+1}, y^1_{m+1})$ and $(x^2_{m+1},$ $ y^2_{m+1})$ as the forward neighboring point of $(x_m, y_m)$ while $(x_m, y_m)$ is the common backward neighboring point of $(x^1_{m+1}, y^1_{m+1})$ and $(x^2_{m+1}, y^2_{m+1})$.
All foreground pixels on the $\overline{S}$ are processed following the same formula and then $\overline{T}_{f,b}$ and $\overline{D}_{f,b}$ can be generated. 
The process is shown in Fig.~\ref{fig:lane_encoder}.

\subsection{Lane decoder with transfer and distance map}

With the predictions of local location and global geometry, we propose a novel lane decoding algorithm to detect all curves in a given image. 

Given the predicted binary segmentation mask $S$, tranfer map $T$ and distance map $D$, we collect all the foreground points of $S$ and use a Point-NMS to get a sparse set of key points $K$. 
Every key point $p \in K$ serves as a start point to recover one global curve.

\noindent\textbf{Step1}: Find the forward transfer vector $T_f(p)$ and forward distance scalar $D_f(p)$ for $p$. 
The moving steps we should extend the neighbors for the forward branch can be defined as $M^{f} = \frac{D_f(p)}{d}$. 
In other words, we can infer the location of the forward end point of $p$ with $D_f(p)$ on the same lane.

Here $d$ is the step length. Then the forward neighbor pixel $p_{i+1}^{f}$ of $p_i^{f}$ can be calculated iteratively by:

\begin{equation}
p_{i+1}^{f} = p_i^{f} + T_f({p_i^{f}}),\  i \in \{0, 1, 2, \cdots, M^{f} - 1\},\ p_0=p.
\label{eq:forward branch}
\end{equation} 

\begin{figure}[t]
    \setlength{\abovecaptionskip}{0.1cm}
    \centering
    \includegraphics[width=1\linewidth]{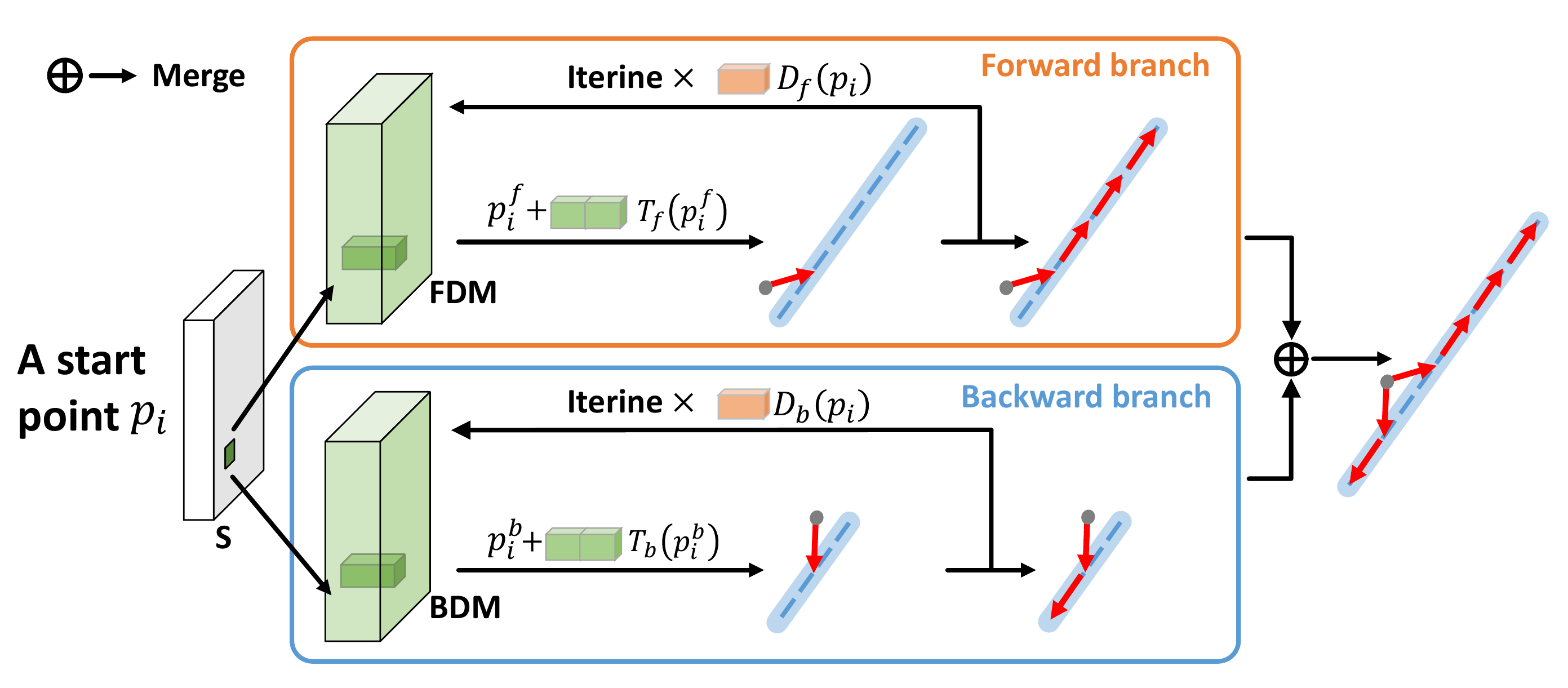}
    \caption{\textbf{The illustration of the lane decoder.} The forward branch predicts the forward part of the lane via forward transfer map $T_f$ and forward distance map $D_f$. The backward part can be decoded from the backward branch similarly.}
    \label{fig:Lane_decoder}
\end{figure}

\noindent The forward branch of the curve can be recovered by connecting $\{p, p_1^{f}, \cdots, p_{M^{f}}^{f}\}$ sequentially. The detail is shown on the top of Fig.~\ref{fig:Lane_decoder}.

\noindent\textbf{Step2}: We calculate the point set $\{p, p_1^{b}, p_2^{b}, \cdots, p_{M^{b}}^{b}\}$ following Eq.~\ref{eq:forward branch} via $T_b$ and $D_b$ and connect them sequentially to recover the backward branch.

\noindent\textbf{Step3}: We then merge the backward and forward curve branches together to get the global curve:

\begin{equation}
\gamma_L=\{p_{M^{b}}^{b}, \cdots, p_2^{b}, p_1^{b}, p, p_1^{f}, p_2^{f}, \cdots, p_{M^{f}}^{f}\}.
\end{equation}
Finally, the non-maximum suppression~\cite{neubeck2006efficient} is performed on all the predicted curves to get the final results.

\subsection{Network architecture}

The overall framework is shown in Fig.~\ref{fig:network archi}. 
SegFormer~\cite{Xie2021SegFormerSA} is utilized as our network backbone, aiming to extract global contextual information and learn the long and thin structures of lanes. 
SegFormer-B0, B1 and B2 are used as small, medium and large backbones in our experiments respectively.
Given an image $I \in R^{H\times W \times 3}$, the segmentation head predicts the binary segmentation mask $S \in R^{H \times W \times 1}$, the transfer head predicts the transfer map $T$ which consists of the forward and backward parts $T_f, T_b$ $ \in \mathbb{R}^{H \times W \times 2}$, and the distance head predicts the distance map $D$ that consists of $D_f, D_b$  $ \in \mathbb{R}^{H \times W \times 1}$.

\subsection{Loss function}
To train our proposed model, we adopt different losses for predictions. 
For the binary segmentation mask, we adopt the OHEM loss~\cite{shrivastava2016training} to train it in order to solve class imbalance problem due to the sparsity of lane segmentation points. The OHEM loss is formulated as follows:

\begin{equation}
L_{seg} = \frac{1}{N_{pos}+N_{neg}}(\sum_{i \in S_{pos}}y_i log(p_i)+\sum_{i \in S_{neg}}(1-y_i)log(1-p_i)).
\end{equation}

where $S_{pos}$ is the set of positive points and $S_{neg}$ is the set of hard negative points which is most likely to be misclassified as positive. 
$N_{pos}$ and $N_{neg}$ denote the number of points in $S_{pos}$ and $S_{neg}$ respectively. The ratio of $N_{neg}$ to $N_{pos}$ is a hyperparmeter $\mu$.
As for the per-pixel transfer and distance maps, we simply adopt the smooth $L_1$ loss, which are denoted as $L_{T}$ and $L_{D}$, to train them. 

\begin{equation}
L_D = \frac{1}{N_{pos}}\sum_{i \in S_{pos}}L_{smooth_{L_1}}(D(p_i), \overline{D}(p_i)),
\end{equation}

\begin{equation}
L_T = \frac{1}{N_{pos}}\sum_{i \in S_{pos}}L_{smooth_{L_1}}(T(p_i), \overline{T}(p_i)).
\end{equation}
In the training phase, the total loss is defined as follows:

\begin{equation}
L_{total} = L_{seg} + L_{T} + L_{D}.
\end{equation}

\section{Experiment}
\subsection{Experimental setting}

\begin{table}[t]
\setlength{\belowcaptionskip}{0.1cm}
\caption{Lane detection datasets.}
\footnotesize
\centering
\begin{tabular}{l|p{1.3cm}<{\centering}p{1.3cm}<{\centering}p{1.3cm}<{\centering}|c}
\toprule[1.3pt]
Dataset &Train &Val. &Test &Road type \\ 
\hline
\hline
CULane~\cite{Pan2018SpatialAD} &88.9k &9.7k &34.7k &urban, highway \\
CurveLanes~\cite{Xu2020CurveLaneNASUL} &100K &20K &30K &urban, highway \\
TuSimple~\cite{tusimple} &3.3k &0.4k &2.8k &highway \\
LLAMAS~\cite{Behrendt2019UnsupervisedLL} &58.3k &20.8k &20.9k &highway \\
\bottomrule[1.3pt]
\end{tabular}
\label{tab:lane dataset}
\end{table}

\noindent\textbf{Dataset.}
We conduct experiments on four widely used lane detection benchmark datasets: CULane~\cite{Pan2018SpatialAD}, 
TuSimple~\cite{tusimple}, LLAMAS~\cite{Behrendt2019UnsupervisedLL} and CurveLanes~\cite{Xu2020CurveLaneNASUL}. CULane consists of 55 hours of videos which comprises nine different scenarios, including normal, crowd, dazzle night, shadow, no line, arrow, curve, cross and night.
The TuSimple dataset is collected with stable lighting conditions on highways. 
LLAMAS is a large lane detection dataset obtained on highway scenes with annotations auto-generated by using high-definition maps.
CurveLanes is a recently proposed benchmark with cases of complex topologies such as Y-shape lanes and dense lanes. The details of four datasets are shown in Tab.~\ref{tab:lane dataset}.

\noindent\textbf{Evaluation metrics.}
For CULane, CurveLanes and LLAMAS, we utilize F1-measure as the evaluation metric. While for TuSimple, accuracy is presented as the official indicator. And we also report the F1-measure for TuSimple.
The calculation method follows the same formula as in CondLaneNet~\cite{Liu2021CondLaneNetAT}.

\noindent\textbf{Implementation details.} 
The small, medium and large versions of our RCLane-Det are used on all four datasets. Except when explicitly indicated, the input resolution is set to $320 \times 800$ during training and testing. 
For all training sessions, we use AdamW optimizer~\cite{loshchilov2018fixing} to train 20 epochs on CULane, CurveLanes and LLAMAS, 70 epochs on TuSimple respectively with a batch size of 32. 
The learning rate is initialized as $6e$-$4$ with a “poly” LR schedule. We set $\eta$ for calculating IOU between lines as 15, the ratio of $N_{neg}$ to $N_{pos}$ $\mu$ as $15$, the minimum distance between any two foreground pixels of in Point-NMS $\tau$ as 2.
We implement our method using the Mindspore~\cite{mindspore} on Ascend 910.

\begin{table*}[t]
\setlength{\belowcaptionskip}{0.1cm}
\caption{State-of-the-art comparison on CULane. Even the small version of our RCLane achieves the state-of-art performance with only 6.3M parameters.}
\centering
\renewcommand\arraystretch{1.17} 
\resizebox{\textwidth}{!}{
\begin{tabular}{l|cccccccccc|ccc}
\hline
Method &Total &Normal &Crowded &Dazzle &Shadow &No line &Arrow &Curve &Cross &Night &Params(M) &FPS\\
\hline
\hline
SCNN~\cite{Pan2018SpatialAD} &71.60 &90.60 &69.70 &58.50 &66.90 &43.40 &84.10 &64.40 &1990 &66.10 &- &7.5\\
CurveLanes-NAS-S~\cite{Xu2020CurveLaneNASUL} &71.40 &88.30 &68.60 &63.20 &68.00 &47.90 &82.50 &66.00 &2817 &66.20 &- &-\\
CurveLanes-NAS-M~\cite{Xu2020CurveLaneNASUL} &73.50 &90.20 &70.50 &65.90 &69.30 &48.80 &85.70 &67.50 &2359 &68.20 &- &-\\
CurveLanes-NAS-L~\cite{Xu2020CurveLaneNASUL} &74.80 &90.70 &72.30 &67.70 &70.10 &49.40 &85.80 &68.40 &1746 &68.90 &- &-\\
LaneATT-S~\cite{Tabelini2020KeepYE} &75.13 &91.17 &72.71 &65.82 &68.03 &49.13 &87.82 &63.75 &1020 &68.58 &13.3 &250\\
LaneATT-M~\cite{Tabelini2020KeepYE} &76.68 &92.14 &75.03 &66.47 &78.15 &49.39 &88.38 &67.72 &1330 &70.72 &23.4 &171\\
LaneATT-L~\cite{Tabelini2020KeepYE} &77.02 &91.74 &76.16 &69.47 &76.31 &50.46 &86.29 &64.05 &1264 &70.81 &18.8 &26\\
LaneAF (DLA-34)~\cite{abualsaud2021laneaf} &77.41 &91.80 &75.61 &71.78 &79.12 &51.38 &86.88 &72.70 &1360 &73.03 &20.2 &-\\
FOLO~\cite{qu2021focus} &78.80 &92.70 &77.80 &75.20 &79.30 &52.10 &89.00 &69.40 &1569 &74.50 &- &-\\
CondLaneNet-S~\cite{Liu2021CondLaneNetAT} &78.14 &92.87 &75.79 &70.72 &80.01 &52.39 &89.37 &72.40 &1364 &73.23 &12.1 &220\\
CondLaneNet-M~\cite{Liu2021CondLaneNetAT}  &78.74 &93.38 &77.14 &71.17 &79.93 &51.85 &89.89 &73.88 &1387 &73.92 &22.2 &152\\
CondLaneNet-L~\cite{Liu2021CondLaneNetAT} &79.48 &93.47 &77.44 &70.93 &80.91 &\textbf{54.13} &90.16 &75.21 &1201 &74.80 &49.9 &58\\
\hline
\textbf{RCLane-S (Ours)} &79.52 &93.41 &77.93 &\textbf{73.32} &80.31 &53.84 &89.04 &75.66 &1298 &74.33 &6.3 &45.6\\
\textbf{RCLane-M (Ours)} &80.03 &93.59 &78.77 &72.44 &\textbf{84.37} &52.77 &90.31&78.39 &\textbf{907} &73.96 &17.2 &43.8\\
\textbf{RCLane-L (Ours)} &\textbf{80.50} &\textbf{94.01} &\textbf{79.13} &72.92 &81.16 &53.94 &\textbf{90.51} &\textbf{79.66} &931 &\textbf{75.10} &30.9 &24.5\\  
\hline
\end{tabular}}
\label{tab:culane result}
\end{table*}
\begin{figure*}[h]
    \centering
    \includegraphics[width=1\linewidth]{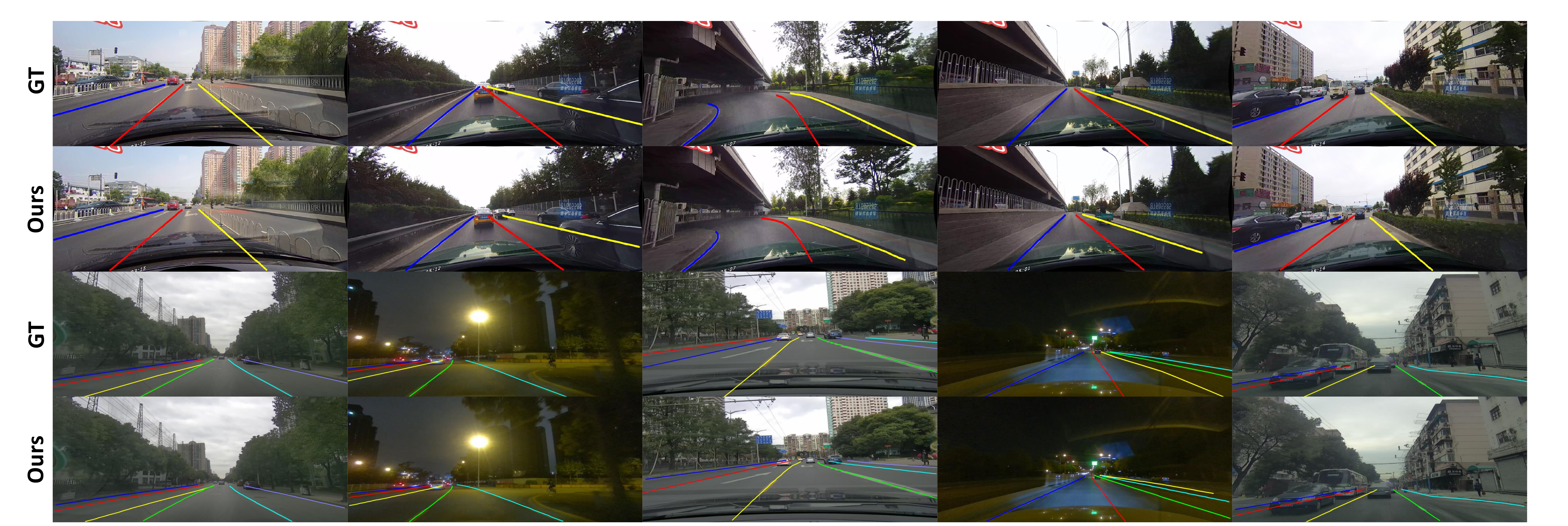}
    \caption{\textbf{Visualization results of our RCLane.} The first two rows are of CULane and the last two rows are of CurveLanes. Our model can detect curve lanes, dense lanes and Y-shape lanes in different scenarios.}
    \label{fig:main results}
\end{figure*}

\subsection{Results}
\textbf{CULane.}
As illustrated in Tab.~\ref{tab:culane result}, RCLane achieves a new state-of-the-art result on the CULane testing set with an 80.50\% F1-measure. 
Compared with the best model as far as we know, CondLaneNet~\cite{Liu2021CondLaneNetAT}, although our method performs better only 1.02\% of F1-measure compared with the best model before CondLaneNet since CULane is a simpler dataset with may straight lines, it has an considerable improvements in crowded and curve scenes, which demonstrates that \textit{Relay Chain} can strengthen local location connectivity through global shape learning for local occlusions and complex lane topologies. 
The visualization result is shown in the first two rows of Fig.~\ref{fig:main results}.

\begin{table}[h]
\setlength{\belowcaptionskip}{0.1cm}
\caption{Performance of different methods on CurveLanes.}
\setlength\tabcolsep{5pt}
\centering
\footnotesize
\begin{tabular}{l|ccc}
\toprule[1.3pt]
Method &F1(\%) &Precision(\%) &Recall(\%)  \\
\hline
\hline
SCNN~\cite{Pan2018SpatialAD}  &65.02 &76.13 &56.74\\
Enet-SAD~\cite{Hou2019LearningLL} &50.31 &63.60 &41.60\\
PointLaneNet~\cite{8813778} &78.47 &86.33 &72.91\\
CurveLane-S~\cite{Xu2020CurveLaneNASUL} &81.12 &93.58 &71.59\\
CurveLane-M~\cite{Xu2020CurveLaneNASUL} &81.80 &93.49 &72.71\\
CurveLane-L~\cite{Xu2020CurveLaneNASUL} &82.29 &91.11 &75.03\\
CondLaneNet-S~\cite{Liu2021CondLaneNetAT} &85.09 &87.75 &82.58\\
CondLaneNet-M~\cite{Liu2021CondLaneNetAT} &85.92 &88.29 &83.68\\
CondLaneNet-L~\cite{Liu2021CondLaneNetAT} &86.10 &88.98 &83.41\\
\hline
\textbf{RCLane-S (Ours)} &90.47 &93.33 & 87.78\\ 
\textbf{RCLane-M (Ours)} &90.96 &93.47 &88.58\\
\textbf{RCLane-L (Ours)} &\textbf{91.43} &\textbf{93.96} &\textbf{89.03}\\   
\bottomrule[1.3pt]
\end{tabular}
\label{tab:curvelane result}
\end{table}
\begin{figure*}[h]
    \setlength{\abovecaptionskip}{0.1cm}
    \centering
    \includegraphics[width=1\linewidth]{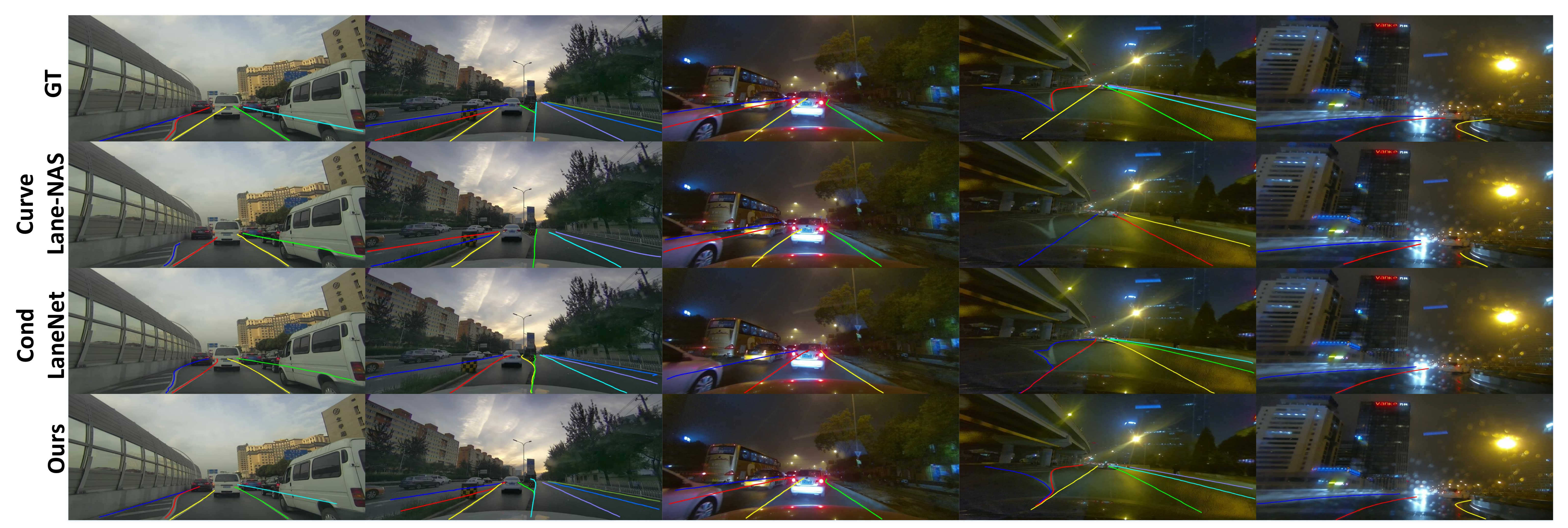}
    \caption{\textbf{Comparison of visualization results} between our RCLane and other lane detection methods. Our method  performs better in different scenarios such as occluded, curved and Y-shape lanes.}
    \label{fig:compare result}
\end{figure*}

\noindent\textbf{CurveLanes.}
CurveLanes~\cite{Xu2020CurveLaneNASUL} is a challenging benchmark with many hard scenarios.
The evaluation results are shown in Tab.~\ref{tab:curvelane result}.
We can see that our largest model (with SegFormer-B2) surpasses CondLaneNet-L by 5.33\% in F1-measure, which is more pronounced than it on CULane. 
Due to that CurveLanes is more complex with Fork-shape, Y-shape and other curve lanes,  improvements both in recall rate and accuracy prove that RCLane has generalization ability on lanes. 
The visualization results is shown in the last two rows of Fig.~\ref{fig:main results}. 
And the qualitative comparison with other methods is shown in Fig.~\ref{fig:compare result}.

\noindent\textbf{TuSimple.}
The results on TuSimple are shown in Tab.~\ref{tab:tusimple result}. 
As Tusimple is a small dataset and scenes are more simple with accurate annotations, the gap between all methods is small.
Moreover, our method also achieves a new state-of-the-art F1 score of 97.64\%. 

\begin{table}[h]
\setlength{\belowcaptionskip}{0.1cm}
\caption{Performance of different methods on TuSimple.}
\footnotesize
\centering
\begin{tabular}{l|cccc}
\toprule[1.3pt]
Method &F1(\%) &Acc(\%) &FP(\%) &FN(\%) \\
\hline
\hline
SCNN~\cite{Pan2018SpatialAD} &95.97 &96.53 &6.17 & \textbf{1.80}\\
PointLaneNet~\cite{8813778} & 95.07 &96.34 &4.67 &5.18\\
LaneATT-ResNet18~\cite{Tabelini2020KeepYE} &96.71 &95.57 &3.56 &3.01\\
LaneATT-ResNet34~\cite{Tabelini2020KeepYE} &96.77 &95.63 &3.53 &2.92\\
LaneATT-ResNet122~\cite{Tabelini2020KeepYE} &96.06 &96.10 &5.64 &2.17\\
CondLaneNet-S~\cite{Liu2021CondLaneNetAT}  &97.01 &95.48 &2.18 &3.80\\
CondLaneNet-M~\cite{Liu2021CondLaneNetAT}  &96.98 &95.37 &2.20 &3.82\\
CondLaneNet-L~\cite{Liu2021CondLaneNetAT}  & 97.24 &96.54 &\textbf{2.01} &3.50\\
LaneAF(DLA-34)~\cite{abualsaud2021laneaf}  &96.49 &95.62 &2.80 &4.18\\
FOLO~\cite{qu2021focus}  &- &\textbf{96.92} &4.47 &2.28\\
\hline
\textbf{RCLane-S (Ours)}  &97.52 &96.49 &2.21 &2.57\\
\textbf{RCLane-M (Ours)} &97.61 &96.51 &2.24 &2.36 \\
\textbf{RCLane-L (Ours)} &\textbf{97.64} &96.58 &2.28 &2.27\\
\bottomrule[1.3pt]
\end{tabular}
\label{tab:tusimple result}
\end{table}

\noindent\textbf{LLAMAS}
LLAMAS~\cite{Behrendt2019UnsupervisedLL} is a new dataset with more than $100K$ images from highway scenarios. The results of our RCLane on LLAMAS is shown in Tab.~\ref{tab:llamas result}. The best result of our method is 96.13\% F1 score with RCLane-L.

\begin{table}[h]
\setlength{\belowcaptionskip}{0.1cm}
\caption{Performance of different methods on LLAMAS.}
\footnotesize
\setlength\tabcolsep{3pt}
\centering
\begin{tabular}{l|ccc}
\toprule[1.3pt]
Method &F1(\%) &Precision(\%) &Recall(\%)  \\
\hline
\hline
PolyLaneNet~\cite{Tabelini2021PolyLaneNetLE} &88.40 &88.87 &87.93 \\
LaneATT-ResNet-18~\cite{Tabelini2020KeepYE} &93.46 &\textbf{96.92} &90.24 \\
LaneATT-ResNet-34~\cite{Tabelini2020KeepYE} &93.74 &96.79 &90.88 \\
LaneATT-ResNet-122~\cite{Tabelini2020KeepYE} &93.54 &96.82 &90.47 \\
LaneAF(DLA-34)~\cite{abualsaud2021laneaf} &96.07 &96.91 &95.26 \\
\hline
\textbf{RCLane-S (Ours)} &96.05 &96.70 &95.42 \\
\textbf{RCLane-M (Ours)} &96.03 &96.62 &95.45 \\  
\textbf{RCLane-L (Ours)} &\textbf{96.13} &96.79 &\textbf{95.48}\\ 
\bottomrule[1.3pt]
\end{tabular}
\label{tab:llamas result}
\end{table}

\subsection{Ablation study}
\textbf{Different modules.} In this section, we perform the ablation study to evaluate the impact of the proposed relay station construction, bilateral predictions and global shape message learning on CurveLanes. 
The results is shown in Tab.~\ref{tab:ablation study1}. 
The first row shows the baseline result, which only uses binary segmentation plus post processing named DBSCAN~\cite{ester1996density} to detect lanes. 
In the second row, the lane is recovered from bottom to top gradually with the guidance of the forward transfer map and forward distance map. While the third row detect lanes from top to bottom. 
In the fourth row, we only use the forward and backward transfer maps to predict the lane. 
And we present our full version of RCLane in the last row, which attains a new state-of-art result 91.43\% on CurveLanes.

Comparing the first two rows, we can see that the proposed relay station construction has greatly improved the performance. 
Then, we add global shape information learning with distance map which can improve the performance from 88.19\% to 91.43\%. 
While we do additional two experiments in the second and third lines, the lane is detected by transfer and distance maps from one-way direction and there is a certain gap with the highest F1-score.
It proves that our bilateral prediction has generalization in depicting topologies of lanes. 
In addition, there exists a gap between the forward  the backward models. 
As the near lanes (the bottom region of the image) are usually occluded by the ego car, the corresponding lane points get low confidence scores from the segmentation results. 
Therefore the starting points are usually outside of the occluded area and the forward counterpart eventually has no chance back to cover the lanes at the bottom of the image.
In contrast, the backward model detects lanes more completely with the help of the distance map when decoding from the top, including the occluded area.
\begin{table}[t]
\footnotesize
\setlength{\belowcaptionskip}{0.1cm}
\caption{Comparison of different components on CurveLanes. The $T_f$, $T_b$, $D_f$, $D_b$ represent the forward transfer map, backward transfer map, forward distance map and backward distance map respectively.}
\centering
\begin{tabular}{p{1.3cm}<{\centering}p{0.9cm}<{\centering}<{\centering}<{\centering}<{\centering}p{0.9cm}<{\centering}<{\centering}<{\centering}p{0.9cm}<{\centering}<{\centering}p{0.9cm}<{\centering}|l}
\toprule[1.3pt]
Baseline &$T_f$ &$T_b$ &$D_f$ &$D_b$ &F1(\%) \\
\hline
\hline
$\surd$ & & & & &$51.22$\\ 
&$\surd$ & &$\surd$ & &$75.06^{\tiny{\textbf{+23.84}}}$\\ 
& &$\surd$ & &$\surd$ &$83.78^{\tiny{\textbf{+32.56}}}$\\ 
&$\surd$ &$\surd$ & & &$88.19^{\tiny{\textbf{+36.97}}}$\\ 
\hline
&$\surd$ &$\surd$ &$\surd$ &$\surd$ &$91.43^{\tiny{\textbf{+40.21}}}$\\ 
\bottomrule[1.3pt]
\end{tabular}
\label{tab:ablation study1}
\end{table}

\noindent\textbf{Comparisons with other methods using the same backbone.}
We additionally use Segformer-B2~\cite{Xie2021SegFormerSA} as backbone to train CondLaneNet~\cite{Liu2021CondLaneNetAT} and LaneAF~\cite{abualsaud2021laneaf} respectively and show their results on Tab.~\ref{tab:backbone result} below. 
Without changing the parameters of their models, our model still outperforms LaneAF and CondLaneNet by a margin on CULane~\cite{Pan2018SpatialAD} dataset due to its superior precision, which demonstrates the high quality of lanes detected by RCLane. 
It further fairly verifies the superiority of our proposed relay chain prediction method, which can process local location and global geometry information simultaneously to improve the capacity of the model. 
\begin{table}[h]
\footnotesize
\setlength{\belowcaptionskip}{0.1cm}
\caption{Comparisons with other methods using the same backbone Segformer-B2.}
\centering
\scalebox{1.0}{
\setlength\tabcolsep{3pt}
\begin{tabular}{l|ccc}
\toprule[1.3pt]
method &Precision(\%) & Recall(\%) & F1(\%)\\
\hline
\hline
LaneAF~\cite{abualsaud2021laneaf} &80.89 & 71.71 & 76.02\\
CondLaneNet~\cite{Liu2021CondLaneNetAT} &82.58 &\textbf{76.01} &79.16\\
\textbf{RCLane (Ours)} &\textbf{88.52} &73.82 &\textbf{80.50}\\
\bottomrule[1.3pt]
\end{tabular}}
\label{tab:backbone result}
\end{table}

\noindent\textbf{Experimental setting for step length $\mathbf{d}$.}
Step length $d$ is the distance between the two neighbors when encoding the lane, which is fixed as 10 for all the experiments in our method. 
LaneATT~\cite{Tabelini2020KeepYE} sets the line width as $30$ for calculating IoU. We set it as $15$ according to the resolution scale initially. 
And $d$ should be at least half of the line width to ensure all foreground points find neighbors sited at the center line.
Thus we set $d$ as 7, 8, 9, 10, 11 and 12 and conduct a series of experiments on CurveLanes~\cite{Xu2020CurveLaneNASUL}. The quantitative results are shown in Tab.~\ref{tab:step length}. 
We find that F1-score decreases as $d$ increases since small step length can describe the local variable shape of lanes more precisely while increasing decoding time. 
10 is chosen as the final setting considering the performance-speed trade-off.
\begin{table}[h]
\setlength{\belowcaptionskip}{0.1cm}
\caption{Ablation study on step length on CurveLanes.}
\footnotesize
\centering
\setlength\tabcolsep{3pt}
\begin{tabular}{c|ccc}
\toprule[1.3pt]
Step length  &Precision(\%) & Recall(\%) & F1(\%)\\
\hline
\hline
7  &94.29 &88.87 &91.50 \\
8  &94.26 &88.84 &91.47\\
9 &94.28 &88.79 &91.45\\
10 &93.96 &89.03 &91.43\\
11 &93.93 &88.97 &91.38\\
12 &94.16 &88.93 &91.31\\
\bottomrule[1.3pt]
\end{tabular}
\label{tab:step length}
\end{table}

\begin{figure}[ht]
    \centering
    \includegraphics[width=1\linewidth]{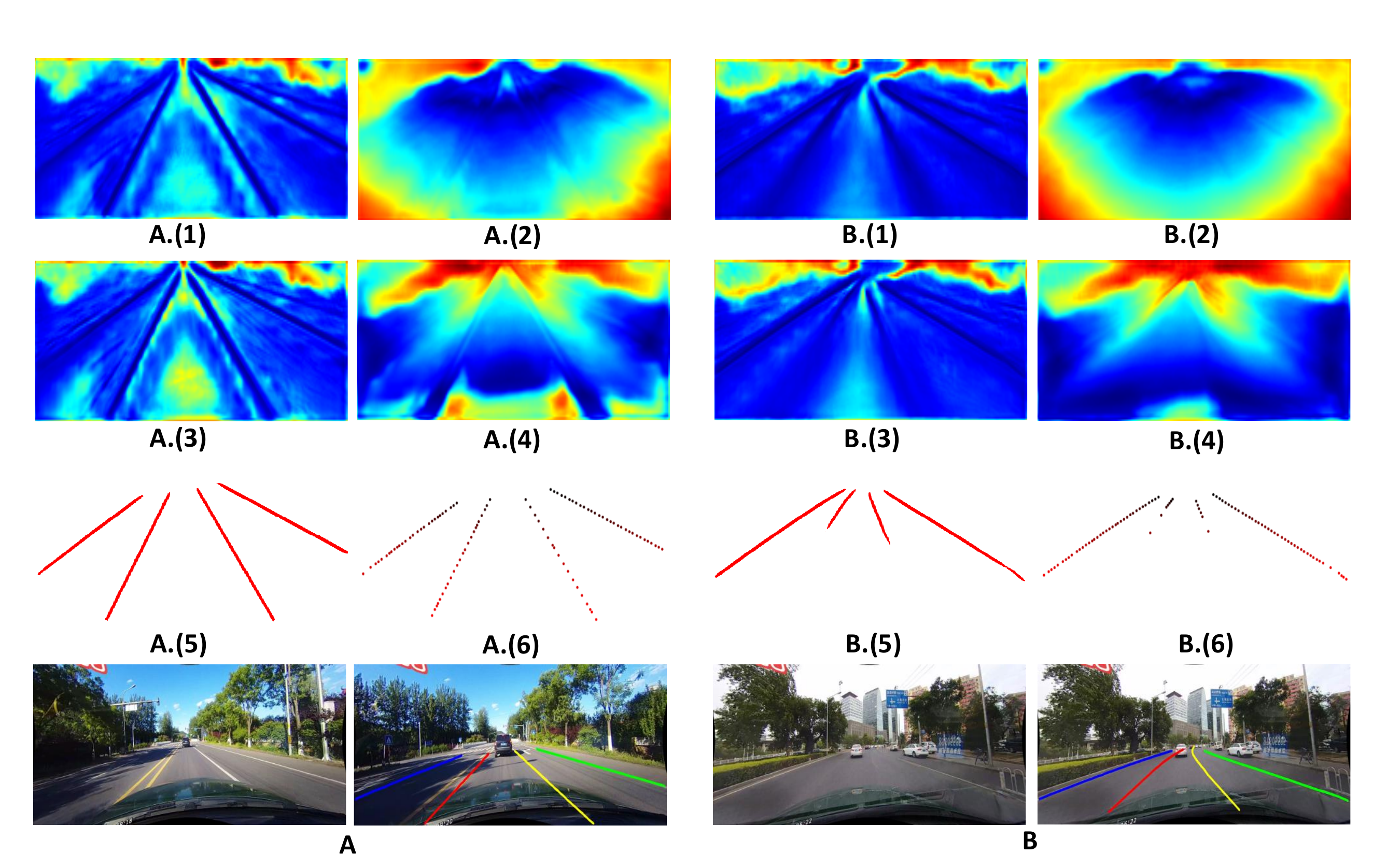}
    \caption{\textbf{Visualization of network outputs.} A.(1, 3) are features of $D_f$ and $D_b$, while A.(2, 4) are features of $T_f$ and $T_b$. A.(5) is the segmentation result and becomes sparse map A.(6) via Point-NMS. B is a harder frame compared to A. }
    \label{fig:feat_vis}
\end{figure}

\noindent\textbf{Local location and global shape message modeling.} In Fig.~\ref{fig:feat_vis} $A.(1,3)$, the transfer map can capture local location information depicting topology of the lane precisely, while the distance map in Fig.~\ref{fig:feat_vis} $A.(2,4)$ models global shape information with large receptive field. 
Furthermore, in some driving scenarios, there occurs loss of lane information due to the disappearance of trace for lanes as is shown in Fig.~\ref{fig:feat_vis}(B). 
However, lanes are still captured faintly in the transfer map with the global shape information learning. 
The results show the robustness of our RCLane with local location and global shape message modeling. 
\section{Conclusion}
In this paper, we have proposed to solve lane detection problem by learning a novel relay chain prediction model.
Compared with existing lane detection methods, our model is able to capture global geometry and local information progressively with the novel relay station construction and global shape message learning. 
Furthermore, bilateral predictions can adapt to hard topologies, such as Fork-shape and Y-shape.
Extensive experiments on four benchmarks including CULane, CurveLanes, Tusimple and LLAMAS demonstrate state-of-the-art performance and generalization ability of our \textit{RCLane}.

\section*{Acknowledgments}
This work was supported in part by 
National Natural Science Foundation of China (Grant No. 6210020439),
Lingang Laboratory (Grant No. LG-QS-202202-07),
Natural Science Foundation of Shanghai (Grant No. 22ZR1407500),
Shanghai Municipal Science and Technology Major Project (Grant No. 2018SHZDZX01 and 2021SHZDZX0103),
Science and Technology Innovation 2030 - Brain Science and Brain-Inspired Intelligence Project (Grant No. 2021ZD0200204),
MindSpore and CAAI-Huawei MindSpore Open Fund.

\bibliographystyle{splncs04}
\bibliography{egbib}

\appendix
\section{Appendix}

\subsection{Attention mechanisms}
We consider the recent attention mechanism is suitable for information processing and do ablation experiments on \textit{CurveLanes}.
Results are shown in Tab.~\ref{tab:attn result}. Adding self-attention~\cite{Vaswani2017AttentionIA} to U-Net~\cite{ronneberger2015u}, which operated on the deepest feature map, makes F1-score increase from  $89.41\%$ and reaches $89.49\%$. 
In the third row, we replace it with axial attention~\cite{wang2020axial} and further improves the performance considering the row-column style attention adapts to the long and thin characteristics of lanes.
Finally, we utilize the efficient transformer-based network SegFormer as backbone and achieve the best result.
From the above results, we can find attention mechanism can help our model focus on local location and global shape information simultaneously.
\begin{table}[h]
\footnotesize
\setlength{\belowcaptionskip}{0.1cm}
\caption{Comparison among different settings of attention on the CurveLanes. The attention module is just added to the feature map of the smallest resolution.}
\centering
\setlength\tabcolsep{3pt}
\begin{tabular}{l|ccc}
\toprule[1.3pt]
Backbone &Precision(\%) & Recall(\%) & F1(\%)\\
\hline
\hline
Unet~\cite{ronneberger2015u} &93.11 & 86.00 &89.41\\
Unet~\cite{ronneberger2015u} + self-attention~\cite{Vaswani2017AttentionIA} &92.95 & 86.27 & $89.49^{\tiny {\textbf{+0.08}}}$\\
Unet~\cite{ronneberger2015u} + axial-attention~\cite{wang2020axial} &92.79 &88.41 &$90.55^{\tiny {\textbf{+1.14}}}$ \\
SegFormer-B2~\cite{Xie2021SegFormerSA} &93.96 &89.03 &$91.43^{\tiny {\textbf{+2.02}}}$\\
\bottomrule[1.3pt]
\end{tabular}
\label{tab:attn result}
\end{table}

\subsection{Generalization} This section aims to verify the generalization capacity of our RCLane following FOLOLane~\cite{qu2021focus}.
We utilize the model trained on the CULane~\cite{Pan2018SpatialAD} training set to evaluate on the TuSimple~\cite{tusimple} test set. 
The results are shown in Tab.~\ref{tab:gene res}. 
Our RCLane surpasses FOLOLane~\cite{qu2021focus} by 2.88\% , with smaller FP and FN indicating that our method is more robust than previous lane detection methods. 
\begin{table}[h]
\setlength{\belowcaptionskip}{0.1cm}
\caption{Evaluation of generalization ability of different methods from CULane training set to TuSimple testing set.}
\centering
\setlength\tabcolsep{3pt}
\begin{tabular}{l|ccc}
\toprule[1.3pt]
Method  & Accuracy(\%) &FP(\%) & FN(\%)\\
\hline
\hline
PINet~\cite{Ko2020KeyPE}&36.31 &48.86 &89.88\\
UFNet~\cite{Qin2020UltraFS}  &65.53 & 56.80 &65.46\\
FOLOLane~\cite{qu2021focus} &84.36 &39.64 &38.41\\
\hline
\textbf{RCLane (Ours)} &\textbf{87.24} &\textbf{22.06} &\textbf{21.56}\\
\bottomrule[1.3pt]
\end{tabular} 
\label{tab:gene res}
\end{table}

\subsection{Visualization results on Tusimple and LLAMAS}
\begin{figure*}[h]
    \centering
    \includegraphics[width=1\linewidth]{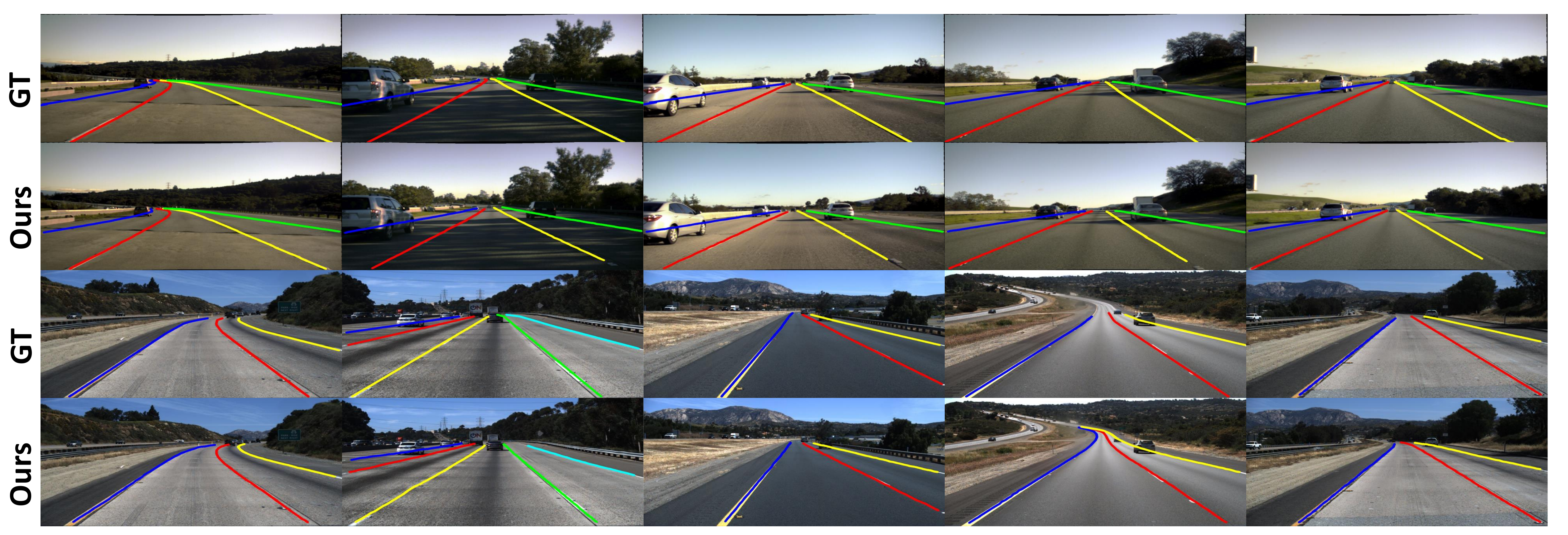}
    \caption{\textbf{Visualization on Tusimple and LLAMAS.} The first two rows are the ground truth and our predictions on LLAMAS and the last two rows are the ground truth and our predictions on Tusimple.}
    \label{fig:tusimple and LLAMAS main}
\end{figure*}

The qualitative results on TuSimple~\cite{tusimple} and LLAMAS~\cite{Behrendt2019UnsupervisedLL} are shown in Fig.~\ref{fig:tusimple and LLAMAS main}. TuSimple and LLAMAS are two benchmarks taken from the highway driving scenarios and are easier compared with CULane~\cite{Pan2018SpatialAD} and CurveLanes~\cite{Xu2020CurveLaneNASUL}. 
In some scenarios such as curve lanes or the far end of lanes, our RCLane even shows better performance than ground truths, as is shown in the last row and forth column in Fig.~\ref{fig:tusimple and LLAMAS main}.

\subsection{Visualizations on complex scenes}
\begin{figure*}[h]
    \setlength{\abovecaptionskip}{0.1cm}
    \centering
    \includegraphics[width=1\linewidth]{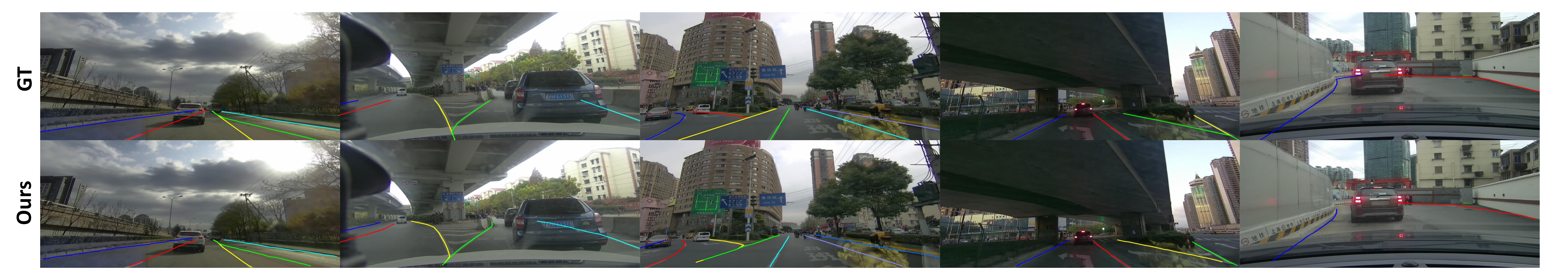}
    \caption{\textbf{Visualization of detection results for complex topologies,} such as fork-shape, Y-shape and nearly horizontal lanes, shows the capacity of RCLane. }
    \label{fig:hard case result}
\end{figure*}

Moreover, we visualize more results of our RCLane on other hard cases on CurveLanes in Fig~\ref{fig:hard case result} and achieve good performance both of Fork-shape lanes and nearly horizontal lanes. 

\subsection{Pseudo code}
\subsubsection{Pseudo code for lane encoder.}
Algorithm·\ref{alg1} details the process of lane encoder, which aims to generate the ground truth of the transfer map and distance map to supervise the training process.

\begin{algorithm} 
\renewcommand{\algorithmicrequire}{\textbf{Input:}}
\renewcommand{\algorithmicensure}{\textbf{Output:}}
\caption{Lane encoder} 
\label{alg1} 
\begin{algorithmic}[1] 
\REQUIRE 
$d$: the step length\\
$\overline{S}$: the foreground point set in segmentation map\\
$L$: the set of lanes in the image $I$\\
$g_1(\cdot)$: the function to compute the distance between a point and a lane\\
$g_2(\cdot)$: the function to compute the distance between two points\\
\ENSURE
\textit{\protect $\overline{T}_f$}: the forward transfer map\\
\textit{\protect $\overline{T}_b$}: the backward transfer map\\
\textit{\protect $\overline{D}_f$}: the forward distance map\\
\textit{\protect $\overline{D}_b$}: the backward distance map\\

\FORALL{$p_i=(x_i, y_i) \in \overline{S}$}
\STATE $l = \mathop{\arg\min}_{\gamma \in L}g_1(p_i, \gamma)$
\IF{$g_1(p_i, l) < d$}
\STATE find the forward and backward end points of $l$:\\
$p_{end}^f = (x_{end}^f, y_{end}^f) = \mathop{\arg\max}_{(x, y) \in l}(y)$,\\
$p_{end}^b = (x_{end}^b, y_{end}^b) = \mathop{\arg\min}_{(x, y) \in l}(y)$
\STATE compute the forward and backward distance scalars:\\

$\overline{D}_f(p_i) = \sqrt{(x_i-x_{end}^{f})^2 + 
(y_i-y_{end}^{f})^2}$,\\
$\overline{D}_b(p_i) = \sqrt{(x_i-x_{end}^{b})^2 + 
(y_i-y_{end}^{b})^2}$

\STATE find the forward and backward point for $p_i$ on lane $l$:\\
$p_i^f=(x^f, y^f) = \mathop{\arg\max}_{ g_2((x, y), p_i) = d}(y)$,\\
$p_i^b=(x^f, y^f) = \mathop{\arg\min}_{ g_2((x, y), p_i) = d}(y)$
\STATE compute the two transfer vectors:\\
$\overline{T}_f(p_i) = (x_i^{f} - x_i, y_i^{f} - y_i)$,\\
$\overline{T}_b(p_i) = (x_i^{b} - x_i, y_i^{b} - y_i)$\\

\ENDIF 
\ENDFOR
\STATE {\textbf{return} $\overline{T}_f$,$\overline{T}_b$,$\overline{D}_f$ and $\overline{D}_b$}

\end{algorithmic} 
\end{algorithm}
\clearpage

\subsubsection{Pseudo code for lane decoder.}
Algorithm~\ref{alg2} introduces the detail process for lane decoder, showing how to recover all possible lanes based on the predicted segmentation map, transfer map and distance map. 
\begin{algorithm} 
\renewcommand{\algorithmicrequire}{\textbf{Input:}}
\renewcommand{\algorithmicensure}{\textbf{Output:}}
\caption{Lane decoder} 
\label{alg2} 
\begin{algorithmic}[1]  
\REQUIRE  $d$ : the step length\\
$r$: the minimal distance between two foreground points in key point map\\
$f(\cdot, r)$: the function to perform Point-NMS on segmentation map to get a sparse key point map\\
$h(\cdot)$: the function to perform IOU-NMS on the predict set of lanes\\
$S$: the predicted segmentation map\\
\textit{T}: the predicted transfer map\\
\textit{D}: the predicted distance map\\

\ENSURE {$L$: the set of lanes in the image $I$}

\STATE key point map $K = f(S, r)$
\STATE initialize $L'$ as an empty set
\FORALL{key point $p \in K$}
\STATE initialize two point sets: $G_1=\{p\}$ and $G_2=\{p\}$
\STATE initialize two points: $p_1=p$ and $p_2=p$
\FOR{$i=1$ to $\frac{D_f(p)}{d}$}
\STATE get the forward transfer vector ${T_f}(p_1)$ from $T_f$
\STATE update $p_1$: $p_1 = p_1 +  {T_f}(p_1)$
\STATE update $G_1$: $G_1 = G_1 \cup \{p_1\}$
\ENDFOR

\FOR{$i=1$ to $\frac{D_b(p)}{d}$}
\STATE get the backward transfer vector ${T_b}(p_2)$ from $T_b$
\STATE update $p_2$: $p_2 = p_2 +  {T_b}(p_2)$
\STATE update $G_2$: $G_2 = G_2 \cup \{p_2\}$
\ENDFOR

\STATE get the lane $l_p = \{G_1\} \cup \{G_2\}$ 
\STATE update $L'$: $L' = L' \cup \{l_p\}$
\ENDFOR
\STATE the final results of predicted lanes $L = h(L')$
\STATE {\textbf{return} the set of lanes $L$}
\end{algorithmic} 
\end{algorithm}

\end{document}